\PassOptionsToPackage{table}{xcolor}
\documentclass[sigconf,9pt]{acmart}
\setcopyright{none}
\renewcommand\footnotetextcopyrightpermission[1]{}
\acmConference[]{}{}{}
\acmBooktitle{}

\usepackage[utf8]{inputenc}
\usepackage[T1]{fontenc}
\usepackage{graphics}
\usepackage{graphicx}
\usepackage{booktabs}
\usepackage{subcaption}
\usepackage{hyperref}
\usepackage{url}
\usepackage{amsmath}
\usepackage{mathtools}
\usepackage{amsthm}
\usepackage{nicefrac}
\usepackage{microtype}
\usepackage{xcolor}
\usepackage{multirow}
\usepackage{colortbl}
\usepackage{enumitem}
\usepackage[super]{nth}
\usepackage[normalem]{ulem}
\usepackage{wrapfig}
\usepackage{threeparttable}
\usepackage{titletoc}
\usepackage{lipsum}

\usepackage{algorithm}
\usepackage{algpseudocode}
\algtext*{EndWhile}
\algtext*{EndIf}
\algtext*{EndFor}
\algtext*{EndProcedure}

\usepackage[capitalize,noabbrev]{cleveref}

\definecolor{mygreen}{RGB}{79,173,91}

\theoremstyle{plain}

\theoremstyle{definition}

\theoremstyle{remark}

\usepackage[textsize=tiny]{todonotes}

\makeatletter
\def\@mkbibcitation{\relax}
\makeatother

\DeclareUnicodeCharacter{FF0C}{,}
\DeclareUnicodeCharacter{FF08}{(}
\DeclareUnicodeCharacter{FF09}{)}
\DeclareUnicodeCharacter{FF5E}{\textasciitilde}

\begin{document}

\title{EXPLORE: \underline{Explo}ration with Guided Sea\underline{r}ch for Analog Topology G\underline{e}neration using Language Models}

\author{Guanglei Zhou\textsuperscript{1}, Chen-Chia Chang\textsuperscript{1}, Yikang Shen\textsuperscript{2}, Jonathan Ku\textsuperscript{1}, Isaac Jacobson\textsuperscript{1}, Jingyu Pan\textsuperscript{1}, Yiran Chen\textsuperscript{1}, Xin Zhang\textsuperscript{3}}
\affiliation{%
  \institution{\textsuperscript{1}Electrical and Computer Engineering, Duke University, Durham, USA}
  \country{}}
\affiliation{%
  \institution{\textsuperscript{2}MIT-IBM Watson AI Lab, Cambridge, USA \quad \textsuperscript{3}IBM T. J. Watson Research Center, Yorktown Heights, USA}
  \country{}}
\affiliation{%
  \institution{\textsuperscript{1}\{guanglei.zhou, chenchia.chang, jonathan.ku, isaac.jacobson, jingyu.pan, yiran.chen\}@duke.edu}
  \country{}}
\affiliation{%
  \institution{\textsuperscript{2}yikang.shen@ibm.com \quad \textsuperscript{3}xzhang@us.ibm.com}
  \country{}}

\renewcommand{\shortauthors}{Zhou et al.}

\begin{abstract}
Automating analog circuit topology design is essential to reduce the extensive manual effort required to meet increasingly diverse and customized application demands. 
Recent advances have applied sequence-to-sequence fine-tuning on pretrained language models to directly generate circuit topologies from user specifications in a single pass. 
However, these one-shot generation methods failed to generate complex circuits due to their exponentially growing search spaces and limited training datasets.
In this paper, we present \textbf{EXPLORE}, a search-enhanced framework that integrates simulator-guided Monte Carlo Tree Search (MCTS) with transformer-based decoding to enable test-time scaling for analog topology generation. 
By leveraging language-model priors and bypassing high-confidence structural tokens, EXPLORE allocates expensive simulator budget primarily toward topology-altering decisions during search.
On a 6-component benchmark at a tight tolerance of 0.01, EXPLORE raises the success rate from 12\% for one-shot generation and 33\% for a sampling-and-filter baseline to 65\%, and lowers MSE by over 20× relative to sampling-and-filter under the same search budget. 
These results establish EXPLORE as the first framework to integrate structured test-time search with LM decoding for analog topology generation, and a practical step toward scaling LLM-driven design automation.
\end{abstract}

\maketitle

\section{Introduction} 
\begin{figure}[t]
  \centering
  \includegraphics[width=0.4\textwidth]{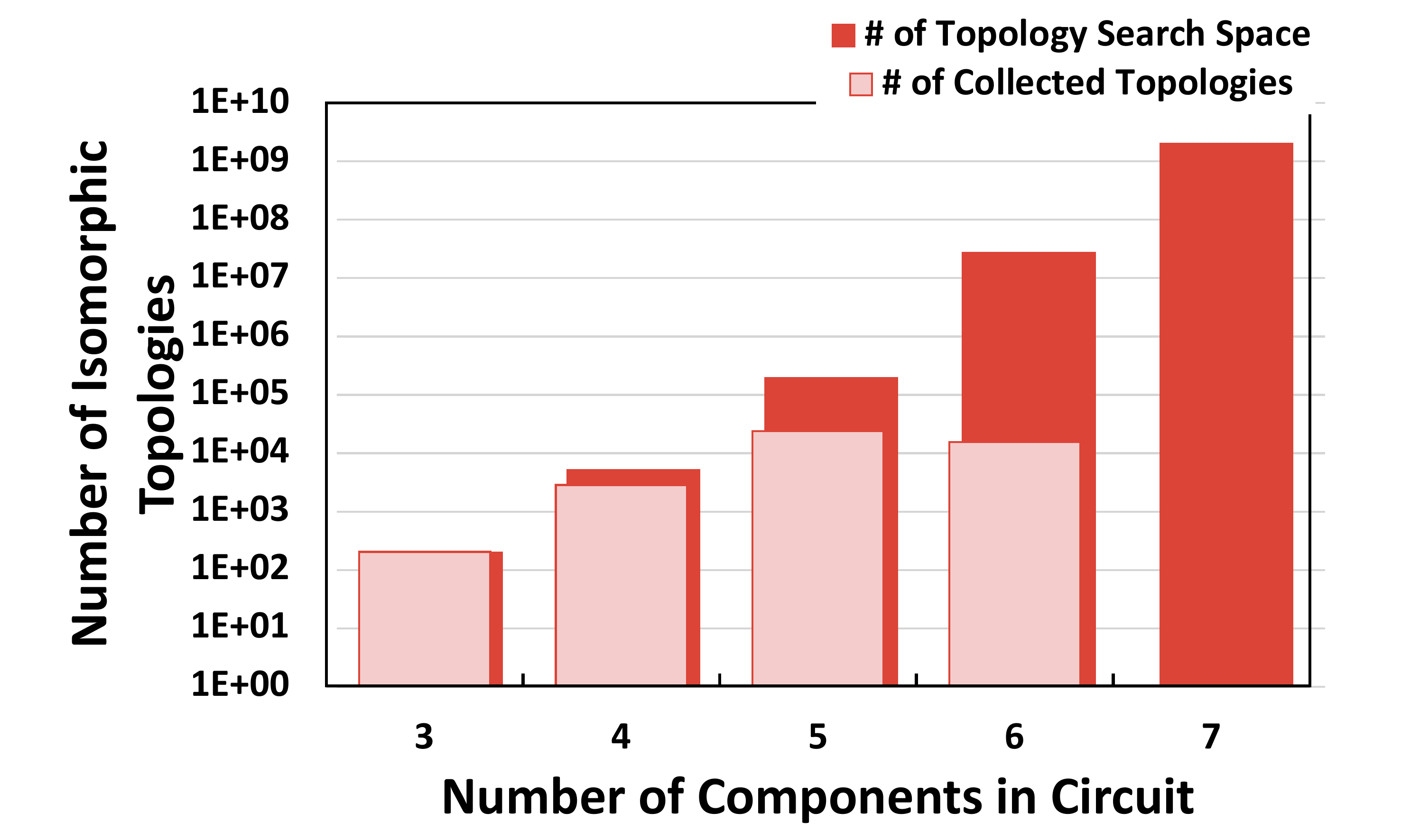}  
\caption{Topology space grows combinatorially with circuit complexity, while collected topologies remain limited, resulting in increasingly sparse design-space coverage~\cite{Analog_Graph_DAC,LAMAGIC}.}
  \label{fig:topo_space}
\end{figure}

Analog circuit topology design sits at the heart of modern electronic systems, enabling everything from efficient power conversion to high-speed signal processing. As device requirements proliferate, varying voltage-conversion ratios, efficiency targets, and performance specifications, the burden on designers to craft bespoke topologies grows heavier. Traditional workflows remain largely manual, demanding extensive domain expertise and hundreds of simulation iterations per new requirement, which in turn prolongs development cycles and delays time-to-market. To meet these challenges, automating the topology design process has become essential: by embedding search and learning methods directly into the design flow, engineers can rapidly explore vast design spaces, reduce iteration counts, and accelerate the creation of optimized analog circuits.

Early works~\cite{AnalogRL_ICCAD, Analog_opamp_TCAD22, Analog_TCAD} leverage reinforcement learning (RL) or Bayesian optimization to discover valid topologies using simulation feedback. These methods reduce evaluation costs and produce functional designs, but suffer from two key limitations: (1) they must restart search or retrain policies for every new specification, and (2) they have only demonstrated success on small, relaxed 3–5 component circuits. Without access to prior knowledge across tasks, such methods remain inefficient and unscalable for practical usage.

LaMAGIC~\cite{LAMAGIC} reframed topology generation as a sequence-to-sequence problem for autoregressive language models, introducing several text-based circuit formulations.
Trained on a corpus of 132k 3---5 component converter topologies, LaMAGIC achieved strong results within this regime. 
However, as Figure~\ref{fig:topo_space} illustrates, the topology search space grows exponentially with component count. 
At six components, a good dataset coverage becomes impractical: with an average simulation time of 9 seconds per topology, enumerating the full design space would require over 2000 CPU-days. 
Consequently, LaMAGIC~\cite{LAMAGIC} struggles to transfer knowledge from 345 components to six, and scaling further to complex circuits with 8–10 components becomes infeasible. 
This gap highlights a fundamental bottleneck: existing approaches, constrained by limited datasets, cannot scale topology generation on their own. Overcoming this limitation requires not only new computing paradigms, such as search-enhanced test-time scaling techniques, but also novel dataset collection strategies that efficiently harvest high-quality training data in larger, more complex circuit spaces.
In this work, we propose \textbf{EXPLORE}, a search-enhanced language model framework for automated analog topology generation. EXPLORE integrates simulator-guided Monte Carlo Tree Search (MCTS) with transformer-based decoding to leverage inference-time compute for generating higher-quality and more scalable analog circuit topologies. Building on text-based circuit formulations, our framework combines pretrained language-model priors with simulator feedback to guide exploration toward valid and high-performance circuit structures. In addition, we construct a large-scale dataset of higher-complexity analog topologies to study the scalability of language-model-based topology generation beyond prior works.

The main contributions of this work are as follows:

\begin{itemize}[nosep,topsep=2pt,leftmargin=*]

\item \textbf{First test-time-scaling framework.}
EXPLORE is the first work to bring test-time scaling to analog
topology generation, along two axes: an LM-guided MCTS as a
test-time decoder, and the same search reused as a model-native
data-collection engine.

\item \textbf{Structured-token filtering.}
We bypass structural tokens in the text-based
circuit formulation via p-filtering ($p=0.99$), skipping 24--48\%
of expansive simulation trials and making scaling to
higher-complexity circuits computationally feasible.

\item \textbf{Superior generation success and query efficiency.} At a tight tolerance of 0.01 on 6-comp, EXPLORE raises the success rate from 12\% for one-shot LaMAGIC and 33\% for its sampling-and-filter variant to 65\%, with over 20× lower MSE than sampling-and-filter under the same search budget and faster convergence than our MCTS-Base ablation within 100 generations.

\item \textbf{Scalability beyond one-shot generation method.}
EXPLORE enables generation at 7, 8, and 9 components, a regime
where one-shot LM decoding collapses to near-zero success and which
prior 3--6 component datasets do not cover.

\end{itemize}

\section{Preliminaries} 
\subsection{Analog topology design}

\begin{figure}[t]{}{}  
  \centering
  \includegraphics[width=0.45\textwidth]{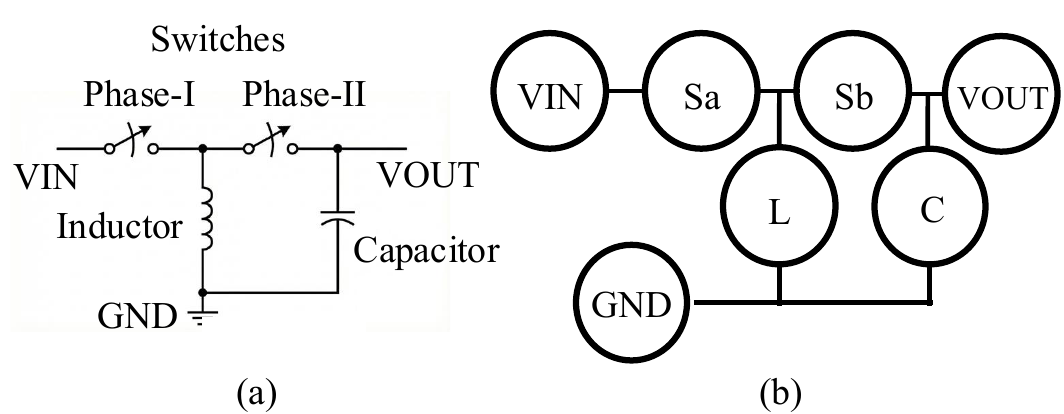}  
  \caption{(a) An example power converter circuit and (b) its corresponding graph representation.} 

  \label{fig:problem_formulation}
\end{figure}

In this work, we address the same problem as LaMAGIC~\cite{LAMAGIC}: generating customized power converters that meet specific voltage conversion ratios and efficiency targets. The \textbf{voltage conversion ratio} is the output-to-input voltage ratio, while power conversion \textbf{efficiency} is the output-to-input power ratio. The \textbf{duty cycle} (a number between 0-1) is a design parameter, which controls the ON time of switches, affecting performance. We use five discrete duty cycles: ${0.1, 0.3, 0.5, 0.7, 0.9}$. We represent circuits as hypergraphs $G$ with vertices $V$ and hyperedges $E$. Vertices include three terminals (input $V_{\text{IN}}$, output $V_{\text{OUT}}$, and ground GND) and four component types (capacitors $C$, inductors $L$, and switches $S_a$, $S_b$). Hyperedges define connections between components and terminals.
Figure~\ref{fig:problem_formulation} shows an example converter with its hypergraph representation.

\textbf{Problem Statement}: Given vertices $V$, target conversion ratio $r$, and efficiency $\eta$, our model generates connections $E$ and selects duty cycle $s$ to create a circuit meeting both performance requirements.

\subsection{Search- and optimization-based methods}

Classical analog topology generation treats the problem as a black-box search over a combinatorial graph space guided by simulator feedback. \cite{AnalogRL_ICCAD} models power converter design as a sequential decision process using a UCT-based RL tree with physics-guided pruning, achieving up to 67\% fewer SPICE calls than genetic or random search. \cite{Analog_opamp_TCAD22} applies deep RL to op-amp synthesis, combining symbolic analysis and memorization to converge to feasible designs within hours but requiring retraining for each circuit class. \cite{Analog_TCAD} pairs a variational autoencoder with Bayesian optimization to identify spec-compliant topologies more efficiently than graph-grammar engines, and \cite{Analog_Graph_DAC} accelerates the same search loop with a graph-transformer SPICE surrogate. More recently, ADO-LLM~\cite{ADO-LLM} replaces a hand-tuned acquisition function with in-context learning over an LLM to seed Bayesian optimization for analog sizing. Across this line of work, each new specification still triggers a fresh search or policy retraining, and the demonstrated regime is small.

\subsection{Language model-based methods}
A recent paradigm shift reframes topology generation as a conditional sequence generation task: given a target specification, the model decodes a circuit description token-by-token. AnalogCoder~\cite{AnalogCoder}, AnalogCoder-Pro~\cite{AnalogCoderPro}, AnalogXpert~\cite{AnalogXpert}, and Artisan~\cite{Opamp_LLM_DAC24_Zengxuan} prompt or fine-tune general or domain LLMs to emit SPICE-style code or operate over curated subcircuit libraries, primarily targeting general circuit \emph{families}. Closest to our setting, LaMAGIC~\cite{LAMAGIC} fine-tunes an encoder--decoder transformer to map a target $(v^\star, \eta^\star)$ to a converter topology in one autoregressive pass. Subsequent works have optimized this paradigm by compressing adjacency representations to $O(|V|)$~\cite{LAMAGIC2,LAMAGIC2TODAES}, and applying RL fine-tuning~\cite{AutoCircuitRL}. CktGNN~\cite{dong2023cktgnn} pairs adjacency with a graph VAE, and AnalogGenie~\cite{AnalogGenie}/AnalogGenie-Lite~\cite{AnalogGenieLite} adopt Eulerian-circuit and device-pin representations. All of these remain \emph{one-shot} generators: they expend their entire model capacity on a single decoded sequence, with no mechanism for spending extra compute when the first attempt is structurally invalid or off-spec. For complex circuit creation that requires iterative refinement and simulator feedback, this one-shot paradigm is highly inefficient. Diverging from all prior work, EXPLORE introduces test-time scaling through a structured search framework guided by simulator feedback. EXPLORE keeps the text-based formulation but reframes generation as a guided search, drawing on the search-augmented LLM decoding literature~\cite{alphacode,RAP,ToT} to spend additional inference-time compute with simulator feedback on tokens where the model is uncertain.

\subsection{Complex power converter datasets: prior corpora and challenges}
\label{sec:dataset_challenges}
The lack of sufficiently large analog circuit datasets continues to hinder the development of AI-based generative methods for automating analog IC design. Prior corpora such as AnalogGenie~\cite{AnalogGenie}, Align~\cite{kunal2019align}, CktGNN~\cite{dong2023cktgnn}, AMSNet~\cite{tao2024amsnet}, AMSnet-KG~\cite{AMSnetKG}, and AICircuit~\cite{AICircuit} prioritize breadth, covering diverse circuit families with only thousands of examples per circuit type. Despite the substantial manual effort required to curate these samples from textbooks and datasheets, the per-type sample sizes remain too small for a model to learn the internal dynamics of complex circuits. In contrast, we prioritize on power converters and explore scaling towards higher-component topologies within a single circuit class.

Scaling a power converter dataset to higher component counts, however, introduces three compounding challenges: (1) the learning task becomes more complex, so substantially more training samples are required; (2) the fraction of invalid or useless topologies (e.g., disconnected graphs or near-zero-efficiency circuits) rises sharply; and (3) the probability that a purely random arrangement yields an efficient converter shrinks rapidly as the search space grows combinatorially. These obstacles motivate the two-stage construction pipeline detailed in Section~\ref{sec:dataset_construction}.

\section{Search-enhanced language model framework}
\label{sec:method}

\begin{figure*}[t]  
  \centering
  \includegraphics[width=0.999\textwidth]{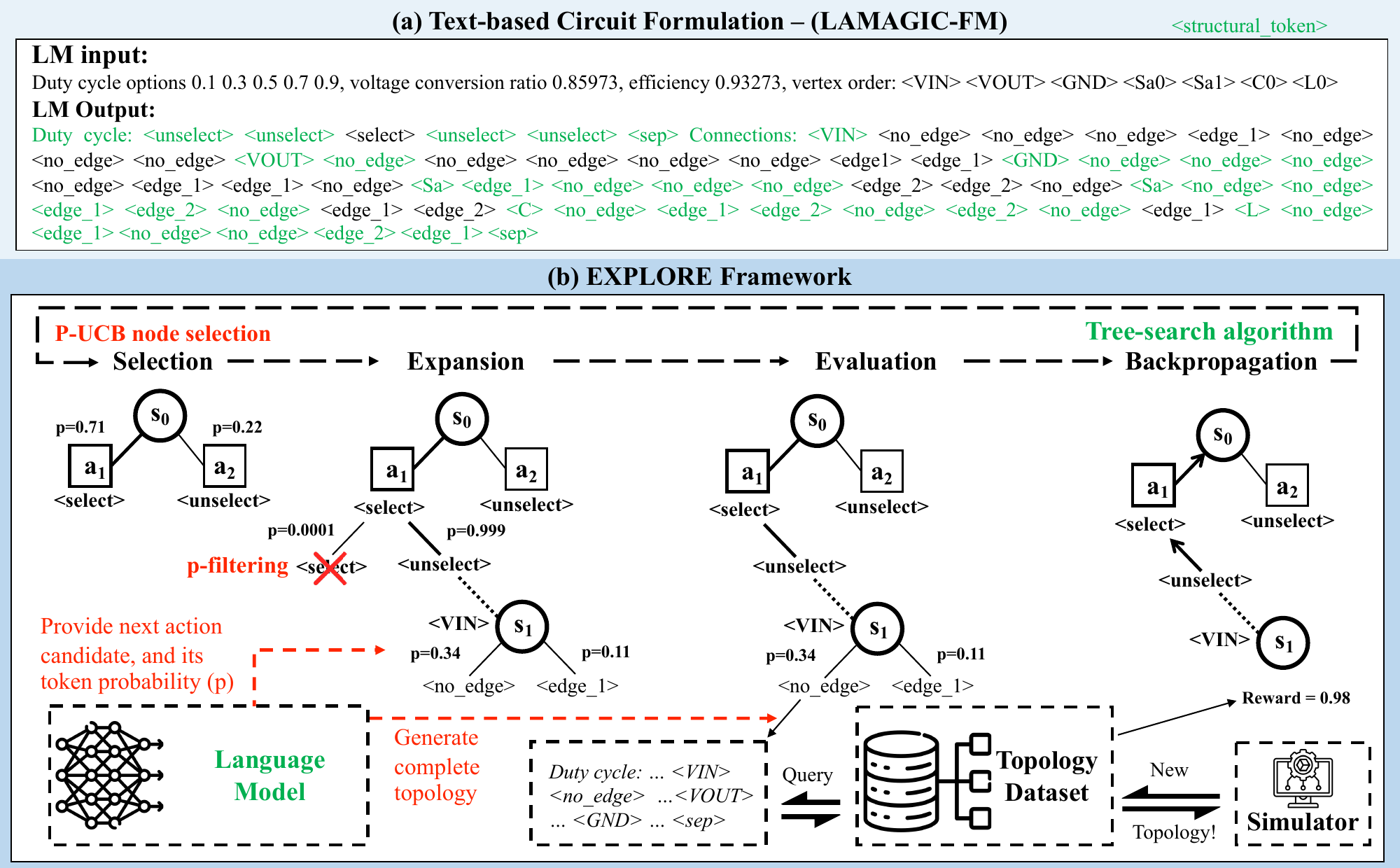}  
  \caption{ (a) A circuit example of float-input adjacency-based matrix formulation for the edge generation task, highlighting its inefficiency due to structural tokens. (b) Illustration of
  the EXPLORE framework pipeline through a step-by-step example of leveraging an MCTS algorithm to guide the Transformer generation for analog circuit topology.}
  \label{fig:framework}
\end{figure*}

We build on LaMAGIC~\cite{LAMAGIC}, whose text-based formulations for circuit generation include the float-input adjacency-matrix formulation (FM), which achieved the best MSE on 6-component circuits. As shown in Figure~\ref{fig:framework}(a), FM represents circuit connections as an adjacency matrix over a hypergraph, with distinct tokens <no edge>, <edge 1>, <edge 2>, and <both edges> indicating the connection type between each vertex pair. A property we exploit below is that more than half of the tokens in an FM sequence are \emph{structural}: they keep the formulation syntactically legal but carry no topological choice (highlighted in Figure~\ref{fig:framework}(a)), which is the opening that p-filtering exploits.

Although our pipeline follows the PUCT-style LM-guided MCTS template used for code generation~\cite{alphacode,PG-TD}, three properties of analog topology generation reshape almost every design choice. (i) \emph{The search target is a graph, not text.} A circuit topology is an undirected (hyper)graph; the FM token stream is just one serialization of its adjacency matrix, so the actual decision space lives over edges between component pairs rather than over the surface tokens that the LM emits. (ii) \emph{Not all tokens are decisions.} As noted above, structural tokens of FM carry no topological choice; in code generation, by contrast, nearly every token changes program semantics. Spending simulator budget on forced tokens is pure waste, which makes p-filtering structurally well-motivated rather than a generic engineering speedup. (iii) \emph{Each evaluation is orders of magnitude more expensive.} A single NGSPICE simulation costs roughly 4 seconds to a few minutes, versus milliseconds for a typical unit test in code-generation settings. Simulator budget, not search depth, is the dominant constraint: LM priors focus expansion on plausible edges, p-filtering removes forced moves, and we beam-complete partial topologies before simulating so every simulator call has a chance of returning useful reward.

As illustrated in Figure~\ref{fig:framework}(b), EXPLORE addresses these constraints with a single MCTS loop guided by transformer LM priors. Each iteration selects a node with PUCT, expands either by auto-committing a high-confidence (forced) token via p-filtering or by spawning top-$k$ children at a genuine decision point, evaluates by beam-completing the partial topology and simulating it in NGSPICE, and backpropagates the resulting reward. The following subsections detail each step.
\subsection{LM-guided topology search}


Algorithm~\ref{alg:MCTS_main} summarizes the procedure; we detail each phase below.

\textbf{Selection.} We apply PUCT-style action priors (in the spirit of AlphaCode~\cite{alphacode} and PG-TD~\cite{PG-TD}) to LM-guided MCTS over circuit tokens: the language-model probability of the child action token enters the UCB exploration term as the action prior $P(a \mid s)$, biasing selection toward likely and under-explored continuations. A tunable exploration constant $c$ controls the trade-off, with higher $c$ encouraging broader search.

\textbf{Expansion.} After selecting a node, we first apply \textbf{p-filtering}: if the language model assigns probability $\geq p$ to its top-1 token (we use $p{=}0.99$), we auto-commit that token and append it to the current node without spawning new children. This pattern is empirically dominated by structural tokens of the FM representation that carry no topological choice. Note that p-filtering is the \emph{opposite} of top-$p$ (nucleus) sampling: nucleus sampling \emph{prunes low-probability} tokens to limit diversity, whereas p-filtering \emph{skips high-probability} tokens that the model is already certain about, so as not to spend simulator budget on forced moves. When no token exceeds the threshold, we apply \textbf{top-$k$ sampling} to generate multiple child nodes, each representing an extended partial topology.

\textbf{Evaluation.} Since partial topologies cannot be directly simulated, we use beam search to complete the sequence from the current node, using a predefined prefix and beam width $b$. The completed topology is then evaluated with NGSPICE to obtain circuit-level performance metrics (output voltage $v$ and conversion efficiency $e$), which are compared to the target $(v^*, e^*)$ using the reward 
\begin{equation}
r \;=\; 0.5 \cdot \max\!\big(0,\, 1 - |v - v^*|\big) \;+\; 0.5 \cdot \max\!\big(0,\, 1 - |e - e^*|\big),
\label{eq:reward}
\end{equation}
clipped to $[0,1]$. The $0.5{/}0.5$ weighting is an unweighted average of voltage and efficiency error; tuning this weighting is left to future work. 

\textbf{Backpropagation} uses the standard PUCT-style mean backup: each ancestor accumulates the leaf reward and exposes $node.value = \tfrac{1}{N}\sum_i r_i$, the running average over the $N$ rollouts that pass through it.
    
    

\begin{algorithm}[H]
\caption{MCTS-based token generation}
\label{alg:MCTS_main}
\begin{algorithmic}[1]
\Require root: initial state; $c$: UCB exploration parameter;\\
        $k$: max children per node; $b$: beam search width;\\
        $p$: threshold for structural token filtering
\Ensure Best sequence from MCTS
\State Initialize tree with root node
\For{$i = 1$ to $max\_rollouts$}
    \State $node \gets root$ \Comment{Selection}
    \While{$node$ has children}
        \State $node \gets$ Select child using UCB
    \EndWhile
    
    \While{len(node.top-p tokens)==1} \Comment{p-filtering}
        \State $node \gets$ CONCAT(node,$next\_tokens$)  \label{alg:line:node_shrinking}
    \EndWhile
    \State $next\_tokens \gets$ top-$k$ tokens \Comment{Expansion}
    
    \ForAll{$token \in next\_tokens$}
        \State Add new child node for $token$ to tree
    \EndFor
    
    \State $sequence \gets$ Perform beam search \Comment{Evaluation}
    \State $r \gets$ Obtain reward via simulation
    \State Backpropagate(node,$r$) \Comment{Backpropagation}
\EndFor
\State \Return sequence with the highest reward
\end{algorithmic}
\end{algorithm}

\subsection{Two-stage dataset construction}
\label{sec:dataset_construction}
To overcome the three scaling challenges identified in Section~\ref{sec:dataset_challenges}, we construct our power-converter corpus in two stages: we first expand the random-enumeration pipeline to bootstrap higher-component data, then close the loop by using the trained model itself, guided by the search procedure above, as a high-quality data generator.

\textbf{Stage 1: expanding the random-enumeration pipeline.} We extend the random topology generator of LaMAGIC~\cite{LAMAGIC} beyond the 3--5 component regime to 6--10 components. We predefine a pool of candidate components (capacitors and inductors at various values, and the two switch types), sample the target number of components, add the three terminals $V_{\text{IN}}$, $V_{\text{OUT}}$, and GND, and draw a random connected topology; each candidate is then simulated in NGSPICE and only valid circuits are retained. Randomizing both the component selection and the connectivity yields topological diversity, but the yield of this pipeline degrades with scale exactly as challenges (2) and (3) predict: for the 7-component set we generated 300{,}000 circuits and only 255{,}707 (85.2\%) were structurally valid, and the vast majority of those exhibit low efficiency. This makes random enumeration alone impractical for building large high-quality datasets at higher component counts.

\textbf{Stage 2: model-native data collection.} Once a model has acquired a basic understanding of the formulation from Stage-1 data, we reuse the trained model together with the EXPLORE framework as a data generator. Because the LM prior biases expansion toward plausible edges and simulator feedback steers the search toward high-efficiency targets, model-native collection attains a far higher useful-sample yield than random enumeration: on 6-component circuits it cuts the fraction of sub-2\%-efficiency topologies from 66.1\% to 18.2\% and raises the fraction above 90\% efficiency from 8.3\% to 23.3\% (analyzed in detail in Appendix~\ref{sec:MCTS_data_collection}). Using this pipeline we assemble, to our knowledge, the largest 6-component converter dataset to date (350k samples), substantially exceeding the 120k 3--5 component public release of LaMAGIC~\cite{LAMAGIC}.

\begin{table*}[t]
\centering
\renewcommand{\arraystretch}{1.2}
\resizebox{0.9\textwidth}{!}{
\scriptsize
\begin{threeparttable}
\begin{tabular}{lcccccc}
\toprule
\multirow{2}{*}{Method}
  & \multicolumn{3}{c}{1\,k training data}
  & \multicolumn{3}{c}{32\,k training data} \\
\cmidrule(lr){2-4} \cmidrule(lr){5-7}
 & Success Rate ($t = 0.01$) & MSE (Voltage) & MSE (Efficiency)
 & Success Rate ($t = 0.01$) & MSE (Voltage) & MSE (Efficiency) \\
\midrule
Greedy            & 0.05 & 0.22   & 0.28   & 0.12 & 0.262  & 0.174  \\
Beam Search       & 0.13 & 0.043  & 0.029  & 0.32 & 0.027  & 0.023  \\
Sampling + Filter & 0.20 & 0.0070 & 0.0081 & 0.33 & 0.007  & 0.008  \\
MCTS-Base         & 0.17 & 0.0154 & 0.0093 & 0.31 & 0.008  & 0.010  \\
\rowcolor{gray!10}
Ours (c=4)        & \textbf{0.52} & \textbf{0.00150} & \textbf{0.00135} & \textbf{0.65} & \textbf{0.00029} & \textbf{0.00019} \\
\bottomrule
\end{tabular}
\end{threeparttable}
}
\vspace{5pt}
\caption{Performance on the 6-comp at threshold \(t=0.01\) for both 1\,k and 32\,k training-data budgets. All methods use up to 100 Transformer generations.}
\label{tab:fm_success}
\end{table*}

\begin{figure*}[t]
    \captionsetup{skip=3pt}
    \captionsetup[subfigure]{skip=5pt}
    \centering
    \begin{subfigure}[c]{0.67\textwidth}
        \centering
        \includegraphics[width=\linewidth]{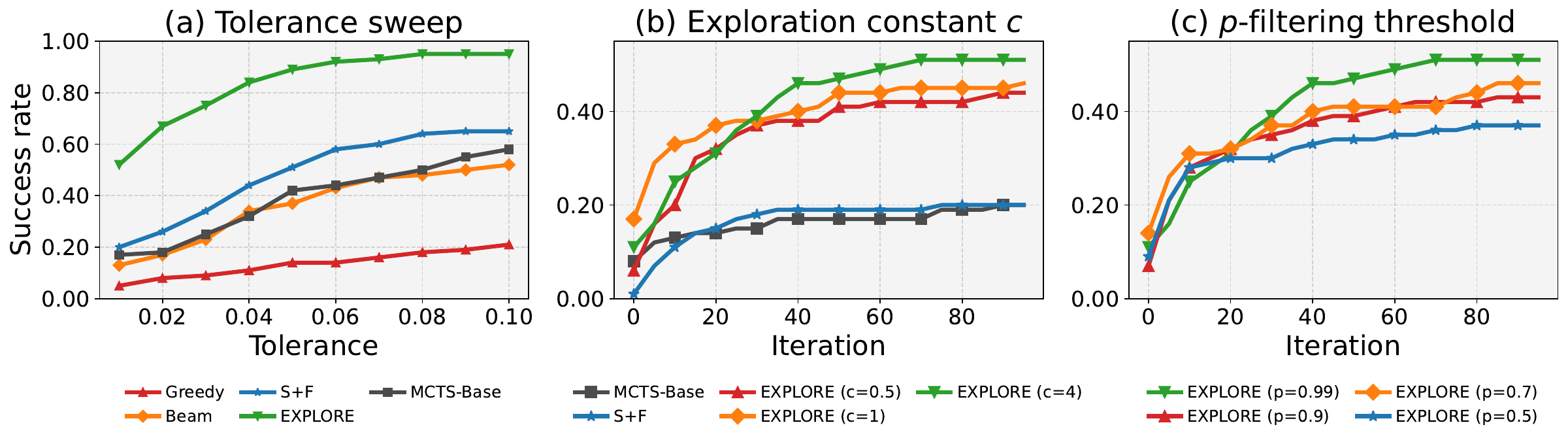}
        \phantomsubcaption\label{fig:exp_tolerance_sweep_hard}
        \phantomsubcaption\label{fig:exp_iter_c_sweep_hard}
        \phantomsubcaption\label{fig:exp_iter_p_sweep}
    \end{subfigure}
    \hfill
    \begin{subfigure}[c]{0.25\textwidth}
        \centering
        \renewcommand{\arraystretch}{1}
        \resizebox{\linewidth}{!}{
        \begin{tabular}{l c}
        \toprule
        \textbf{Metric} & \textbf{Avg Time} \\
        \midrule
        GPU gen. & 0.19s \\
        Sim (no cache) & 10.18s \\
        Sim (cached)   & 4.10s\\
        \midrule
        \textbf{Total time ($n=100$)} & \textbf{9.22h} \\
        \bottomrule
        \end{tabular}
        }
        \caption{Runtime illustration.}
        \label{tab:mcts_runtime_fig}
    \end{subfigure}
\caption{\textbf{Comparison of success rates and runtime analysis on the 6-comp benchmark (FM-1k).}
(\subref{fig:exp_tolerance_sweep_hard}): final success rate under varying error tolerances.
(\subref{fig:exp_iter_c_sweep_hard}): success rate over iterations at $t{=}0.01$ under different exploration constants $c \in \{0.5,1,4\}$.
(\subref{fig:exp_iter_p_sweep}): success rate over iterations at $t{=}0.01$ under different $p$-filtering thresholds $p \in \{0.5,0.7,0.9,0.99\}$ at fixed $c=4$.
(\subref{tab:mcts_runtime_fig})~Runtime breakdown and total evaluation time for EXPLORE.}
    \label{fig:exp_passrate_runtime}
\end{figure*}

\section{Experimental results} 

\subsection{Experiment setup}
\label{sec:exp_setup}
\textbf{Baseline algorithms.} We compare EXPLORE with four decoding baselines. \textbf{Greedy} is the one-shot generation used in LaMAGIC. \textbf{Beam Search} uses Transformer beam search (beam size 20) without simulator feedback. \textbf{Sampling and filtering (S+F)} samples a set of topologies (temperature 1.2, top-$k$ with $k{=}3$ to avoid invalid tokens), simulates each to measure $v_{\text{out}}$ and efficiency, and returns the candidate closest to the target; this mirrors AlphaCode~\cite{alphacode} and a baseline in PG-TD~\cite{PG-TD}. \textbf{MCTS-Base} is our MCTS variant that uses UCB for node selection but ignores LLM token probabilities, treating all equal-count children uniformly, thereby isolating the benefit of LLM-guided priors.

\textbf{Models and datasets.}
Large amount of data for circuits with a higher number of components can be difficult to obtain. To reuse existing knowledge,  similar to the LaMAGIC's setting, we extend models trained with 3,4,5-components to be finetuned with 1k and 32k 6-component circuits and leverage our decoding algorithm to evaluate.
We follow LaMAGIC~\cite{LAMAGIC} to use an encoder-decoder transformer structure with Flan-T5-base pretrained weights. We add a shared linear layer to replace the word embedding layer for numeric inputs. We train the model via conditional generation to learn the mapping between input-output pairs. Model trained with $n$ samples using FM formulation is denoted as FM-$n$. As the full 7k LaMAGIC validation set is costly and dominated by low-performing circuits, our primary benchmark is \textbf{6-comp}: 100 high-efficiency samples spanning various conversion ratios. To confirm the gains are not a selection artifact, we also evaluate \textbf{6-comp-random-100}, 100 samples drawn uniformly from the original set (Appendix~\ref{sec:random500}).

\textbf{Evaluation metrics.}
Our primary metric is success rate: the percentage of generated circuits meeting the preset target $(v^*, e^*)$. For each target and search budget $n$, the method generates up to $n$ candidates and retains the best. A circuit succeeds if its simulated conversion ratio and efficiency $(v, e)$, obtained via NGSPICE~\cite{nenzi2011ngspice}, both fall within tolerance $t$ of the target:
\[
  |v - v^*| \le t
  \quad\text{and}\quad
  |e - e^*| \le t .
\]
We report two variants: (1) \textbf{tolerance-based success rate}, with $t \in \{0.01,0.02,\dots,0.10\}$, and (2) \textbf{iteration-wise success rate}, which tracks success under strict tolerance ($t=0.01$) as the number of generated candidates increases. Candidates that fail to compile or simulate count as failures. We additionally report mean squared errors (MSEs) for the voltage conversion ratio and efficiency to quantify deviation among valid circuits.

\begin{table*}[t]
\centering
\renewcommand{\arraystretch}{1.2}
\resizebox{0.93\textwidth}{!}{
\scriptsize
\begin{tabular}{lcccccccccccc}
\toprule
\multirow{2}{*}{Method}
  & \multicolumn{4}{c}{Success Rate ($t = 0.01$)}
  & \multicolumn{4}{c}{MSE (Voltage)}
  & \multicolumn{4}{c}{MSE (Efficiency)} \\
\cmidrule(lr){2-5} \cmidrule(lr){6-9}\cmidrule(lr){10-13}
 & 7-comp & 8-comp & 9-comp & 10-comp & 7-comp & 8-comp & 9-comp & 10-comp & 7-comp & 8-comp & 9-comp & 10-comp \\
\midrule
Greedy           & 0.02 & 0.02 & 0.04 & 0.00  & 0.763   & 0.904  & 0.906 & 1.019 & 0.338 & 0.348& 0.377 & 0.591\\
Sampling + Filter   & 0.03 & 0.09 & 0.09 & 0.00 & 0.269 & 0.372 & 0.188 & 0.928 & 0.231 & 0.190 & 0.094 & 0.476\\
MCTS-Base    & 0.09 & 0.09 & 0.11 & 0.00 & 0.115 & 0.154 & 0.481 & 0.917 & 0.045 & 0.044 & 0.124 & 0.475\\
\rowcolor{gray!10}
Ours (c=4)       & \textbf{0.21}  & \textbf{0.26} & \textbf{0.16} & \textbf{0.00} & \textbf{0.125}& \textbf{0.009}& \textbf{0.046}& \textbf{0.838} & \textbf{0.024} & \textbf{0.008} & \textbf{0.027} & \textbf{0.432}\\
\bottomrule
\end{tabular}
}
\vspace{5pt}
\caption{Performance at threshold \(t=0.01\) with varying number of components(7 → 10). All methods use up to 100 Transformer generations.}
\label{tab:789}
\end{table*}

\begin{figure*}[]
    \centering
    \includegraphics[width=0.93\textwidth]{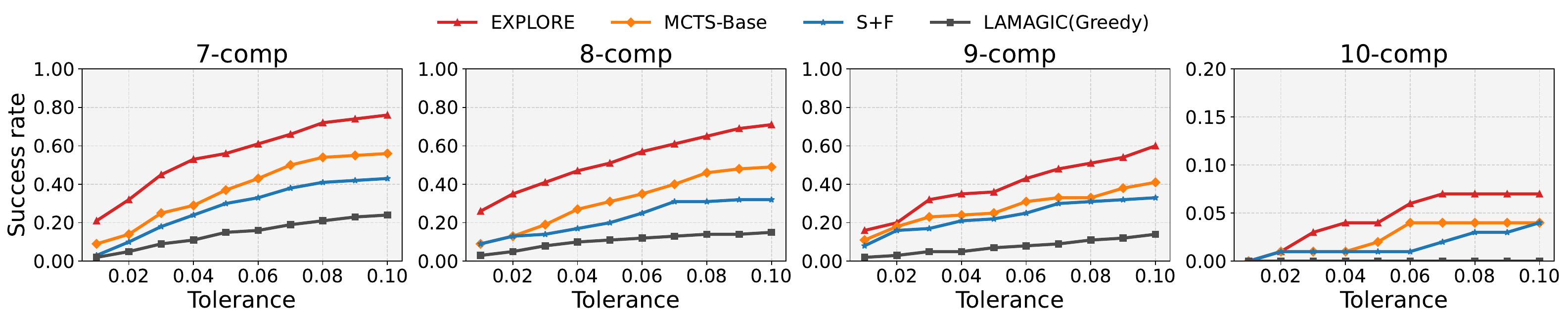}
    \caption{Comparison of generation results on validation sets with varying number of components (7 → 10), for Greedy, Sampling + Filtering (S+F), MCTS-Base, and our EXPLORE  methods.}
    \label{fig:789}
\end{figure*}

\subsection{Generation results on 6-component circuit}
\textbf{Comparison with baselines.}
For a fair comparison, we evaluate the best topology found by the different decoding algorithms when they use the same number of Transformer generations. All methods are evaluated under the same success criterion on 6-comp.

Results are shown in Table~\ref{tab:fm_success}. Our method consistently outperforms all the other baselines on the 6-comp validation set for various tolerance thresholds. Overall, these results confirm that our algorithm indeed generates better topologies for the target voltage and efficiency. As shown in Table~\ref{tab:fm_success}, EXPLORE largely outperforms the one-shot generation methods, reaching a success rate of 0.65 (Ours) versus 0.33 (S+F) when using the FM-32k model at a tight tolerance of 0.01 on 6-comp. S+F uses the same number of transformer generations but is overall outperformed by EXPLORE. Comparing with MCTS-Base, this confirms that the LLM probability guidance is crucial in the node selection stage. A runtime breakdown for our framework is also provided in Figure~\ref{tab:mcts_runtime_fig}.

To illustrate the exploration efficiency of various decoding methods, Figure~\ref{fig:exp_iter_c_sweep_hard} plots the strict-tolerance success rate (t=0.01) as a function of iteration count. For EXPLORE, we fix the top-k sampling budget at k=3 with a single beam (b=1) and sweep the P-UCB exploration constant c. We observe that c =1 yields faster gains in the very early iterations, while a larger c (c=4) drives more exploration and ultimately achieves the highest success. In contrast, without the LLM token probability guidance, MCTS Baselines explores less effectively and converges to sub-optimal results with other baseline sampling and filtering.

We further sweep the p-filtering threshold $p$ at fixed $c{=}4$ (Figure~\ref{fig:exp_iter_p_sweep}). Since p-filtering auto-commits tokens whose model probability already exceeds $p$, a smaller $p$ skips more tokens and saves more budget per iteration. The sweep reveals a clear trade-off: aggressive filtering ($p{=}0.7$) produces the fastest gains in the first few iterations, but skipping tokens the model is less confident about discards genuine decision content and plateaus lower, with the most aggressive setting ($p{=}0.5$) capping at a success rate of 0.37. Setting $p{=}0.99$ auto-commits only near-certain structural tokens, preserving search quality and reaching the highest final success (0.51) while still skipping 24--48\% of token expansions.

\subsection{Generation results on 7--10 component circuit}
\label{sec:789}
A central claim of this work is that test-time search scales more gracefully than one-shot decoding as circuit complexity grows. To test this, we retrain the baseline with 4K samples (1K each from 6--9 components) and build 7/8/9/10-comp validation sets; details are in Appendix~\ref{sec:model_training}.
As shown in Figure~\ref{fig:789}, all methods degrade as components increase from 7 to 10, but ours consistently leads across the 7/8/9/10 benchmarks. Structured search (EXPLORE and MCTS-Base) clearly outperforms the S+F approaches, underscoring its importance for graph-generation tasks such as circuit topology design. Quantitatively (Table~\ref{tab:789}), EXPLORE achieves 10$\times$/13$\times$/4$\times$ success-rate gains over the greedy baseline on 7/8/9-comp, with up to 100$\times$ lower voltage MSE.

\textbf{Unseen 10-component circuits.}
\label{sec:10-comp}
The rightmost panel of Figure~\ref{fig:789} reports a 10-component set entirely outside the 6--9 training range. As expected, few valid circuits are generated, yet two points stand out: (1) even without pre-training, EXPLORE still produces higher-complexity circuits, indicating a scalability trend across component counts absent from training; and (2) its relative advantage persists, remaining the top method in this untrained regime.

Overall, these results support our claim that test-time search scales more effectively than baseline strategies as complexity grows, making it well suited to higher-component and future large-scale designs. We stress that they reflect a \emph{scalability trend} rather than practical readiness, as absolute success rates remain modest beyond 8 components. Further results extending EXPLORE to other analog types (op-amps, bandgap references) via AnalogGenie, and a discussion of MCTS as a data-collection method, are deferred to Appendix~\ref{sec:other_analog_results} and~\ref{sec:MCTS_data_collection}.

\section{Conclusion}
\label{sec:conclusion}
In this work, we propose EXPLORE, a search-enhanced language model framework for analog circuit topology generation. EXPLORE tightly integrates simulator-guided Monte Carlo Tree Search (MCTS) with transformer-based decoding, capitalizing on both the learned priors of pretrained models and the dynamic feedback of circuit simulators to navigate vast design spaces. We adapt MCTS to the text-based adjacency-matrix formulation via PUCT-style action priors from the language model, top-$k$ expansion, and p-filtering that auto-commits high-confidence tokens so simulator budget concentrates on genuine decision points.

Extensive experiments across varying data regimes and difficulty settings confirm that EXPLORE achieves higher success rates and lower error metrics than prior RL-based search methods and language model decoding strategies such as sampling and filtering. We further demonstrate its utility as a high-quality data collection engine for scaling analog datasets in low-coverage regimes. Future work includes discovering more efficient circuits, generalizing to more complex analog designs, and developing search-friendly formulations that reduce structural token overhead.



\newpage

\begin{acks}
This work was supported in part by NSF grant No. 2112562.
\end{acks}

\bibliographystyle{ACM-Reference-Format}
\bibliography{reference}

@inproceedings{LAMAGIC,
  author = {Chang, Chen-Chia and Shen, Yikang and Fan, Shaoze and Li, Jing and Zhang, Shun and Cao, Ningyuan and Chen, Yiran and Zhang, Xin},
  title = {{LaMAGIC}: Language-Model-based Topology Generation for Analog Integrated Circuits},
  booktitle = {Proceedings of the 41st International Conference on Machine Learning},
  series = {ICML'24},
  articleno = {241},
  numpages = {10},
  location = {Vienna, Austria},
  publisher = {JMLR.org},
  year = {2024},
  url = {https://proceedings.mlr.press/v235/chang24c.html}
}

@inproceedings{LAMAGIC2,
  author = {Chang, Chen-Chia and Lin, Wan-Hsuan and Shen, Yikang and Chen, Yiran and Zhang, Xin},
  title = {{LaMAGIC2}: Advanced Circuit Formulations for Language Model-Based Analog Topology Generation},
  booktitle = {Proceedings of the 42nd International Conference on Machine Learning},
  series = {ICML'25},
  location = {Vancouver, Canada},
  publisher = {JMLR.org},
  year = {2025},
  url = {https://openreview.net/forum?id=Y0zXGw0GUk}
}

@article{LAMAGIC2TODAES,
  author = {Chang, Chen-Chia and Lin, Wan-Hsuan and Shen, Yikang and Zhou, Guanglei and Chen, Yiran and Zhang, Xin},
  title = {{LaMAGIC}: Advanced Circuit Formulations for Language-Model-based Topology Generation for Analog Integrated Circuits},
  journal = {ACM Transactions on Design Automation of Electronic Systems},
  volume = {31},
  number = {5},
  pages = {1--21},
  publisher = {Association for Computing Machinery},
  address = {New York, NY, USA},
  month = apr,
  year = {2026},
  doi = {10.1145/3799428}
}

@inproceedings{AnalogCoder,
  author = {Lai, Yao and Lee, Sungyoung and Chen, Guojin and Poddar, Souradip and Hu, Mengkang and Pan, David Z. and Luo, Ping},
  title = {{AnalogCoder}: Analog Circuit Design via Training-Free Code Generation},
  booktitle = {Proceedings of the AAAI Conference on Artificial Intelligence},
  volume = {39},
  number = {1},
  pages = {379--387},
  year = {2025},
  doi = {10.1609/aaai.v39i1.32016}
}

@inproceedings{AnalogRL_ICCAD,
  author = {Fan, Shaoze and Cao, Ningyuan and Zhang, Shun and Li, Jing and Guo, Xiaoxiao and Zhang, Xin},
  title = {From Specification to Topology: Automatic Power Converter Design via Reinforcement Learning},
  booktitle = {2021 IEEE/ACM International Conference On Computer Aided Design (ICCAD)},
  pages = {1--9},
  year = {2021},
  doi = {10.1109/ICCAD51958.2021.9643552}
}

@article{Analog_TCAD,
  author = {Lu, Jialin and Lei, Liangbo and Huang, Jiangli and Yang, Fan and Shang, Li and Zeng, Xuan},
  title = {Automatic Op-Amp Generation From Specification to Layout},
  journal = {IEEE Transactions on Computer-Aided Design of Integrated Circuits and Systems},
  volume = {42},
  number = {12},
  pages = {4378--4390},
  year = {2023},
  doi = {10.1109/TCAD.2023.3296374}
}

@inproceedings{Analog_Graph_DAC,
  author = {Fan, Shaoze and Lu, Haoshu and Zhang, Shun and Cao, Ningyuan and Zhang, Xin and Li, Jing},
  title = {Graph-Transformer-based Surrogate Model for Accelerated Converter Circuit Topology Design},
  booktitle = {Proceedings of the 61st ACM/IEEE Design Automation Conference},
  series = {DAC '24},
  articleno = {172},
  numpages = {6},
  location = {San Francisco, CA, USA},
  publisher = {Association for Computing Machinery},
  address = {New York, NY, USA},
  year = {2024},
  doi = {10.1145/3649329.3656258}
}

@article{Analog_opamp_TCAD22,
  author = {Zhao, Zhenxin and Zhang, Lihong},
  title = {Analog Integrated Circuit Topology Synthesis With Deep Reinforcement Learning},
  journal = {IEEE Transactions on Computer-Aided Design of Integrated Circuits and Systems},
  volume = {41},
  number = {12},
  pages = {5138--5151},
  year = {2022},
  doi = {10.1109/TCAD.2022.3153437}
}

@inproceedings{Opamp_LLM_DAC24_Zengxuan,
  author = {Chen, Zihao and Huang, Jiangli and Liu, Yiting and Yang, Fan and Shang, Li and Zhou, Dian and Zeng, Xuan},
  title = {Artisan: Automated Operational Amplifier Design via Domain-specific Large Language Model},
  booktitle = {Proceedings of the 61st ACM/IEEE Design Automation Conference},
  series = {DAC '24},
  articleno = {39},
  numpages = {6},
  location = {San Francisco, CA, USA},
  publisher = {Association for Computing Machinery},
  address = {New York, NY, USA},
  year = {2024},
  doi = {10.1145/3649329.3655903}
}

@inproceedings{UCB,
  author = {Kocsis, Levente and Szepesv{\'a}ri, Csaba},
  title = {Bandit Based Monte-Carlo Planning},
  booktitle = {Machine Learning: ECML 2006},
  editor = {F{\"u}rnkranz, Johannes and Scheffer, Tobias and Spiliopoulou, Myra},
  pages = {282--293},
  publisher = {Springer Berlin Heidelberg},
  address = {Berlin, Heidelberg},
  year = {2006},
  doi = {10.1007/11871842_29}
}

@manual{nenzi2011ngspice,
  author = {Nenzi, Paolo and Vogt, Holger},
  title = {Ngspice Users Manual Version 23},
  year = {2011},
  url = {https://pkgs.fedoraproject.org/repo/extras/ngspice/ngspice23-manual.pdf/eb0d68eb463a41a0571757a00a5b9f9d/ngspice23-manual.pdf},
  note = {Accessed: 2023}
}

@inproceedings{kunal2019align,
  author = {Kunal, Kishor and Madhusudan, Meghna and Sharma, Arvind K and Xu, Wenbin and Burns, Steven M and Harjani, Ramesh and Hu, Jiang and Kirkpatrick, Desmond A and Sapatnekar, Sachin S},
  title = {{ALIGN}: Open-source Analog Layout Automation from the Ground Up},
  booktitle = {Proceedings of the 56th Annual Design Automation Conference 2019},
  pages = {1--4},
  year = {2019},
  doi = {10.1145/3316781.3323471}
}

@inproceedings{dong2023cktgnn,
  author = {Dong, Zehao and Cao, Weidong and Zhang, Muhan and Tao, Dacheng and Chen, Yixin and Zhang, Xuan},
  title = {{CktGNN}: Circuit Graph Neural Network for Electronic Design Automation},
  booktitle = {International Conference on Learning Representations (ICLR)},
  year = {2023},
  url = {https://openreview.net/forum?id=NE2911Kq1sp}
}

@inproceedings{tao2024amsnet,
  author = {Tao, Zhuofu and Shi, Yichen and Huo, Yiru and Ye, Rui and Li, Zonghang and Huang, Li and Wu, Chen and Bai, Na and Yu, Zhiping and Lin, Ting-Jung and others},
  title = {{AMSNet}: Netlist Dataset for {AMS} Circuits},
  booktitle = {2024 IEEE LLM Aided Design Workshop (LAD)},
  pages = {1--5},
  organization = {IEEE},
  year = {2024},
  doi = {10.1109/LAD62341.2024.10691781}
}

@article{alphacode,
  author = {Li, Yujia and Choi, David and Chung, Junyoung and Kushman, Nate and Schrittwieser, Julian and Leblond, R{\'e}mi and Eccles, Tom and Keeling, James and Gimeno, Felix and Dal Lago, Agustin and Hubert, Thomas and Choy, Peter and de Masson d'Autume, Cyprien and Babuschkin, Igor and Chen, Xinyun and Huang, Po-Sen and Welbl, Johannes and Gowal, Sven and Cherepanov, Alexey and Molloy, James and Mankowitz, Daniel J. and Sutherland Robson, Esme and Kohli, Pushmeet and de Freitas, Nando and Kavukcuoglu, Koray and Vinyals, Oriol},
  title = {Competition-level Code Generation with {AlphaCode}},
  journal = {Science},
  volume = {378},
  number = {6624},
  pages = {1092--1097},
  publisher = {American Association for the Advancement of Science (AAAS)},
  month = dec,
  year = {2022},
  doi = {10.1126/science.abq1158}
}

@inproceedings{AnalogXpert,
  author = {Zhang, Haoyi and Sun, Shizhao and Lin, Yibo and Wang, Runsheng and Bian, Jiang},
  title = {{AnalogXpert}: Automating Analog Topology Synthesis by Incorporating Circuit Design Expertise into Large Language Models},
  booktitle = {2025 International Symposium of Electronics Design Automation (ISEDA)},
  pages = {772--777},
  year = {2025},
  doi = {10.1109/ISEDA65950.2025.11100627}
}

@inproceedings{AnalogGenieLite,
  author = {Gao, Jian and Cao, Weidong and Zhang, Xuan},
  title = {{AnalogGenie-Lite}: Enhancing Scalability and Precision in Circuit Topology Discovery through Lightweight Graph Modeling},
  booktitle = {International Conference on Machine Learning (ICML)},
  year = {2025},
  url = {https://openreview.net/forum?id=KRk0WTII0I}
}

@article{AnalogCoderPro,
  author = {Lai, Yao and Poddar, Souradip and Lee, Sungyoung and Chen, Guojin and Hu, Mengkang and Yu, Bei and Luo, Ping and Pan, David Z.},
  title = {{AnalogCoder-Pro}: Unifying Analog Circuit Generation and Optimization via Multi-modal {LLMs}},
  journal = {IEEE Transactions on Computer-Aided Design of Integrated Circuits and Systems (TCAD)},
  year = {2026},
  doi = {10.1109/TCAD.2026.3673493}
}

@inproceedings{AutoCircuitRL,
  author = {Vijayaraghavan, Prashanth and Shi, Luyao and Degan, Ehsan and Mukherjee, Vandana and Zhang, Xin},
  title = {{AutoCircuit-RL}: Reinforcement Learning-Driven {LLM} for Automated Circuit Topology Generation},
  booktitle = {International Conference on Machine Learning (ICML)},
  year = {2025},
  url = {https://openreview.net/forum?id=NvYwrQbzOb}
}

@inproceedings{ADO-LLM,
  author = {Yin, Yuxuan and Wang, Yu and Xu, Boxun and Li, Peng},
  title = {{ADO-LLM}: Analog Design {Bayesian} Optimization with In-Context Learning of Large Language Models},
  booktitle = {International Conference on Computer-Aided Design (ICCAD)},
  year = {2024},
  doi = {10.1145/3676536.3676816}
}

@inproceedings{AICircuit,
  author = {Mehradfar, Asal and Zhao, Xuzhe and Niu, Yue and Babakniya, Sara and Alesheikh, Mahdi and Aghasi, Hamidreza and Avestimehr, Salman},
  title = {{AICircuit}: A Multi-Level Dataset and Benchmark for {AI}-Driven Analog Integrated Circuit Design},
  booktitle = {Machine Learning and the Physical Sciences Workshop, NeurIPS},
  year = {2024},
  url = {https://arxiv.org/abs/2407.18272}
}

@article{AMSnetKG,
  author = {Shi, Yichen and Tao, Zhuofu and Gao, Yuhao and Zhou, Tianjia and Chang, Cheng and Wang, Yaxing and Chen, Bingyu and Zhang, Genhao and Liu, Alvin and Yu, Zhiping and Lin, Ting-Jung and He, Lei},
  title = {{AMSnet-KG}: A Netlist Dataset for {LLM}-based {AMS} Circuit Auto-Design Using Knowledge Graph {RAG}},
  journal = {ACM Transactions on Design Automation of Electronic Systems (TODAES)},
  volume = {30},
  number = {6},
  articleno = {94},
  year = {2025},
  doi = {10.1145/3736166}
}

@inproceedings{RAP,
  author = {Hao, Shibo and Gu, Yi and Ma, Haodi and Hong, Joshua Jiahua and Wang, Zhen and Wang, Daisy Zhe and Hu, Zhiting},
  title = {Reasoning with Language Model is Planning with World Model},
  booktitle = {Conference on Empirical Methods in Natural Language Processing (EMNLP)},
  pages = {8154--8173},
  year = {2023},
  doi = {10.18653/v1/2023.emnlp-main.507}
}

@inproceedings{ToT,
  author = {Yao, Shunyu and Yu, Dian and Zhao, Jeffrey and Shafran, Izhak and Griffiths, Thomas L. and Cao, Yuan and Narasimhan, Karthik},
  title = {Tree of Thoughts: Deliberate Problem Solving with Large Language Models},
  booktitle = {Advances in Neural Information Processing Systems (NeurIPS)},
  year = {2023},
  url = {https://openreview.net/forum?id=5Xc1ecxO1h}
}

@inproceedings{AnalogGenie,
  author = {Gao, Jian and Cao, Weidong and Yang, Junyi and Zhang, Xuan},
  title = {{AnalogGenie}: A Generative Engine for Automatic Discovery of Analog Circuit Topologies},
  booktitle = {International Conference on Learning Representations (ICLR)},
  year = {2025},
  url = {https://openreview.net/forum?id=jCPak79Kev}
}

@inproceedings{PG-TD,
  author = {Zhang, Shun and Chen, Zhenfang and Shen, Yikang and Ding, Mingyu and Tenenbaum, Joshua B. and Gan, Chuang},
  title = {Planning with Large Language Models for Code Generation},
  booktitle = {International Conference on Learning Representations (ICLR)},
  year = {2023},
  url = {https://openreview.net/forum?id=Lr8cOOtYbfL}
}



\newpage
\appendix

\startcontents[appendix]



\newpage
\section*{Appendix}
\section{Further Explanation of methodologies}
\subsection{P-UCB node selection}
\label{sec:P-UCB}
To guide exploration during tree traversal, we extend the standard Upper Confidence Bound (UCB)~\cite{UCB}  strategy by incorporating the token probability predicted by the language model. The P-UCB score for each child node is computed as:
$$
exploration = \sqrt{\frac{log(node.visit\_count)}{child.visit\_count}}
$$
$$
score = exploitation + c \times exploration * child.token\_probability
$$
where $node.visit\_count$ records the number of times the current (parent) node has been visited, $exploitation$ is $child.value$, and language-model probability are assigned at the parent to the action token that produced this child, i.e., the action prior $P(a\mid s)$ in classical PUCT terminology. The hyperparameter $c$ controls the trade-off between exploration and exploitation; in our experiments we sweep $c \in \{0.5, 1, 4\}$. The selection step picks the child with the highest score; the resulting selection, expansion, evaluation, and backpropagation steps are described in Section~\ref{sec:method}.

\section{Additional experimental setup and results}
\subsection{Model training details and compute resources}
\label{sec:model_training}
The \texttt{Flan-T5-base} model consists of 12 transformer layers in both the encoder and decoder. Each layer includes key and value projections with a dimensionality of 64, a feed-forward network with a hidden size of 2048, and employs 12 attention heads. Overall, the model contains approximately 248 million parameters.

To adapt the tokenizer for our specific application, we add custom tokens to its vocabulary. For the FM task, the following tokens are introduced: \texttt{<sep>}, \texttt{<duty 0.1>}, \texttt{<duty 0.3>}, \texttt{<duty 0.5>}, \texttt{<duty 0.7>}, \texttt{<duty 0.9>}, \texttt{VIN}, \texttt{VOUT}, \texttt{GND}, \texttt{Sa}, \texttt{Sb}, \texttt{C}, \texttt{L}, \texttt{<no edge>}, \texttt{<edge 1>}, \texttt{<edge 2>}, and \texttt{<both edges>}.

Training is conducted on a machine equipped with eight NVIDIA A5000 GPUs. The language model is trained over 30 epochs using the AdamW optimizer with an initial learning rate of $3 \times 10^{-4}$. A cosine learning rate schedule is applied with 300 warmup steps. The batch size is set to 128, L2 regularization is applied with a strength of $10^{-5}$, and the dropout rate is set to 0.1. 

\subsection{Evaluation on Other Analog Component Types}
\label{sec:exp_setup_other}
\label{sec:other_analog_results}
\begin{figure}[h]
    \centering
    \includegraphics[width=0.45\textwidth]{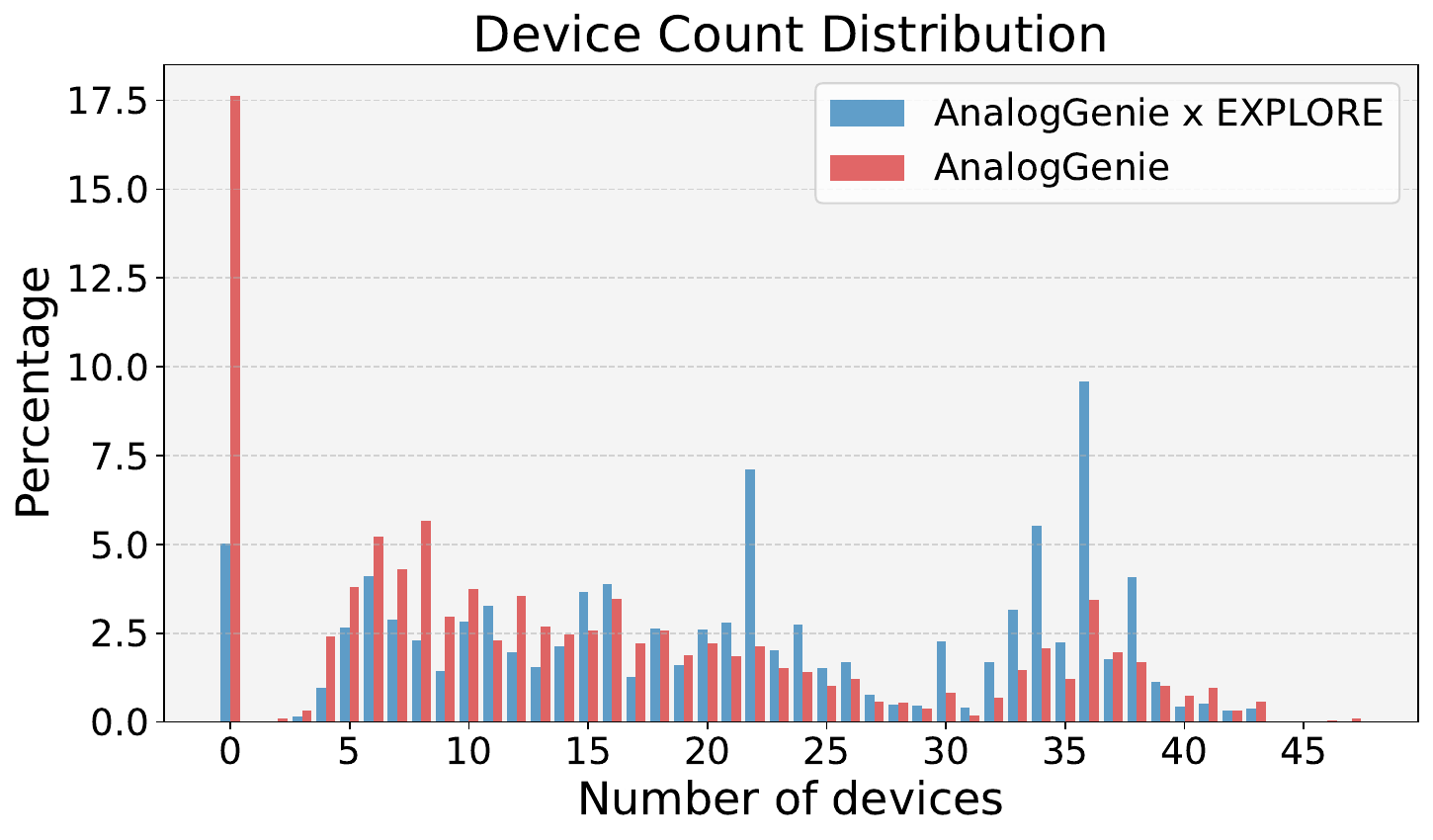}
    \caption{Device count distribution for the first 10000 generated circuits produced by AnalogGenie and AnalogGenie x EXPLORE. Samples with zero devices correspond to invalid generations, including circuits with shorted connections or incomplete pin definitions.}
    \label{fig:AnalogGenie}
\end{figure}

\textbf{Experimental Setup.}
We follow the open-source release of AnalogGenie, using its publicly available model checkpoint on HuggingFace to faithfully reproduce its analog circuit generation behavior. A generated circuit is considered invalid if any device has incomplete pin connectivity (e.g., an NMOS missing one or more of S/D/G/B), or if an electrical short is detected, such as a direct connection between VDD and VSS. We apply our framework using AnalogGenie’s original text-based formulation and conduct experiments under the same generation setting. The reward function is defined as a combination of circuit legality and device count, encouraging valid topologies while discouraging unnecessarily complex designs.

\textbf{Generation Results.}
To evaluate the generality of the proposed formulation, we conduct an ablation study on AnalogGenie for generating additional analog components, including op-amps and bandgap references. Both AnalogGenie and AnalogGenie × EXPLORE are evaluated under the same generation budget (10000 samples) for a fair comparison. As shown in Fig.~\ref{fig:AnalogGenie}, AnalogGenie produces a large fraction of zero-device samples, corresponding to invalid circuits caused by pin shorts or incomplete connections. In contrast, AnalogGenie × EXPLORE substantially reduces the invalid generation rate and shifts the distribution toward higher device counts. This indicates improved topology exploration and more stable generation behavior, demonstrating that EXPLORE generalizes effectively to more complex analog circuit families.

\subsection{MCTS as an effective data collection method}
\label{sec:MCTS_data_collection}
\begin{figure}[h]  
  \centering
  \includegraphics[width=0.5\textwidth]{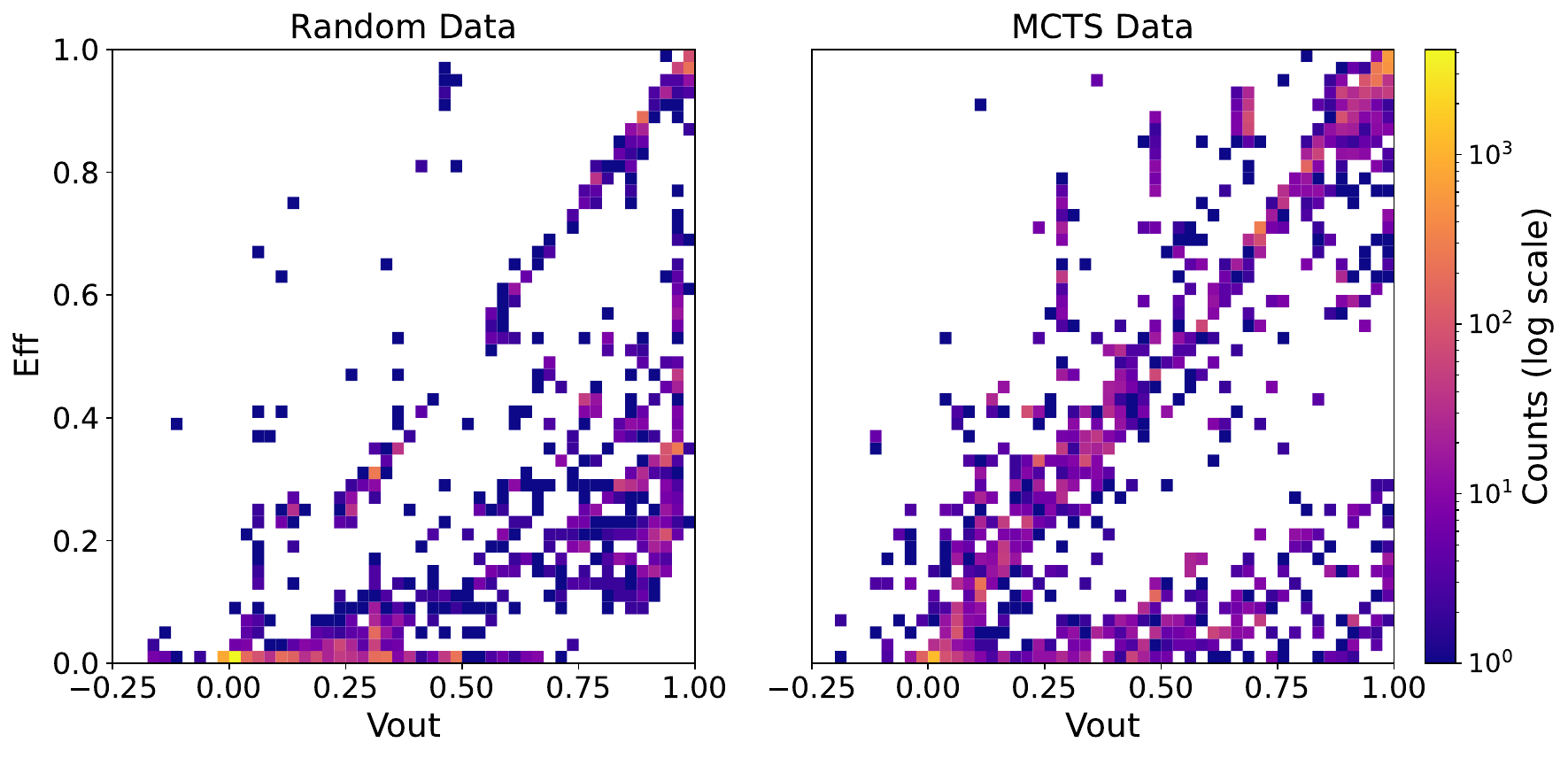}  
  \caption{The Vout vs efficiency distribution of our model collected dataset vs. random connection generated dataset. }
  \label{fig:7_comp_distribution}
\end{figure}
MCTS combined with generative models represents a powerful yet often overlooked approach for high-quality data collection in complex design spaces.
Traditional random generation methods become exponentially ineffective as design complexity grows, with the vast topology search space severely diminishing the probability of discovering valid, high-performance configurations.
Our empirical analysis quantifies this limitation: in a random generation of 10,000 6-component circuits, 66.13\% exhibited efficiency below 2\%, rendering them practically unusable for both application and training purposes. In contrast, our MCTS-based approach significantly mitigates this inefficiency problem, reducing the proportion of low-performing circuits to just 18.2\%. More importantly, our method substantially enhances the discovery of high-quality designs, generating 23.27\% of circuits with efficiency exceeding 90\%-nearly three times higher than the 8.3\% achieved through random generation. The complete efficiency distribution illustrated in Figure~\ref{fig:7_comp_distribution} demonstrates this substantial quality difference.
This shows that beyond immediate circuit applications, our approach can facilitate an efficient mechanism for collecting a high-quality dataset for automatic discovery of unconventional topology and enable further training for the language models.

\subsection{Unbiased Random Evaluation on 6-component}
\label{sec:random500}
The 6-comp subset is mined for high-efficiency targets, which could in principle favor search-based methods. To rule out this selection bias, we evaluate on samples drawn uniformly at random from the full 7k LaMAGIC validation set.

\begin{table}[h]
\centering
\renewcommand{\arraystretch}{1.2}
\resizebox{0.48\textwidth}{!}{
\scriptsize
\begin{threeparttable}
\begin{tabular}{lcccc}
\toprule
Method & \# Train Data & Success Rate ($t = 0.01$) & MSE (Voltage) & MSE (Efficiency) \\
\midrule
Greedy            & 1 000  & 0.21 & 0.33   & 0.16   \\
Beam Search       & 1 000  & 0.41 & 0.033  & 0.022  \\
Sampling + Filter & 1 000  & 0.70 & 0.0216 & 0.0053 \\
MCTS-Base         & 1 000  & 0.62 & 0.0044 & 0.0016 \\
\rowcolor{gray!10}
Ours (c=4)        & 1 000  & \textbf{0.83} & \textbf{0.00016} & \textbf{0.00057} \\
\midrule
Greedy            & 32 000 & 0.37 & 0.195  & 0.165  \\
Beam Search       & 32 000 & 0.51 & 0.025  & 0.013  \\
Sampling + Filter & 32 000 & 0.70 & 0.022  & 0.005  \\
MCTS-Base         & 32 000 & 0.67 & 0.029  & 0.005  \\
\rowcolor{gray!10}
Ours (c=4)        & 32 000 & \textbf{0.84} & \textbf{0.00006} & \textbf{0.00003} \\
\bottomrule
\end{tabular}
\end{threeparttable}
}
\vspace{5pt}
\caption{Performance on the \textbf{6-comp-random-100} split (100 samples drawn uniformly at random from the 7k LaMAGIC validation set) at threshold \(t=0.01\) for both 1\,k and 32\,k training-data budgets. All methods use up to 100 Transformer generations. EXPLORE retains the top success rate and lowest MSE, showing the main-table gain is not an artifact of the 6-comp selection.}
\label{tab:fm_success_random}
\end{table}

We first report \textbf{6-comp-random-100}, a 100-sample uniform draw evaluated under the same metrics as the main 6-comp table (Table~\ref{tab:fm_success_random}). EXPLORE remains the top method at both training budgets, reaching 0.84 success at $t{=}0.01$ with the FM-32k model versus 0.70 for S+F, with the lowest voltage and efficiency MSE.

\begin{table}[h]
\centering
\renewcommand{\arraystretch}{1.2}
\resizebox{0.48\textwidth}{!}{
\scriptsize
\begin{threeparttable}
\begin{tabular}{lccc}
\toprule
Method & Success Rate ($t = 0.01$) & MSE (Voltage) & MSE (Efficiency) \\
\midrule
Greedy            & 0.24 & 0.27    & 0.23    \\
Sampling + Filter & 0.48 & 0.19    & 0.15    \\
MCTS-Base         & 0.59 & 0.0026  & 0.0026  \\
\rowcolor{gray!10}
Ours (c=4)        & \textbf{0.73} & \textbf{0.00063} & \textbf{0.00038} \\
\bottomrule
\end{tabular}
\end{threeparttable}
}
\vspace{5pt}
\caption{Performance on the \textbf{6-comp-random-500} split (500 samples drawn uniformly at random from the 7k LaMAGIC validation set) at threshold \(t=0.01\). Greedy is taken as the first per-target sample from the S+F generations. EXPLORE retains the top success rate and lowest MSE, confirming the main-table gain is not an artifact of the 6-comp selection.}
\label{tab:q3_random500_joint}
\end{table}

To confirm the trend at larger scale, we further evaluate \textbf{6-comp-random-500} (Table~\ref{tab:q3_random500_joint}). EXPLORE is the top method at $t{=}0.01$ (0.73 success vs.\ 0.59 for MCTS-Base, 0.48 for S+F, and 0.24 for Greedy) and also attains the lowest voltage and efficiency MSE by more than an order of magnitude, confirming that the gain reported in Sec.~\ref{sec:exp_setup} is not an artifact of the hard-subset construction. To keep the table comparable to the 6-comp setting without separately rerunning argmax decoding on the random subset, the Greedy row is populated from the first per-target sample of the S+F generations.

\subsection{Effect of LLM Probability Guidance}
\begin{figure}[h]{} 
\centering
\includegraphics[width=0.35\textwidth]{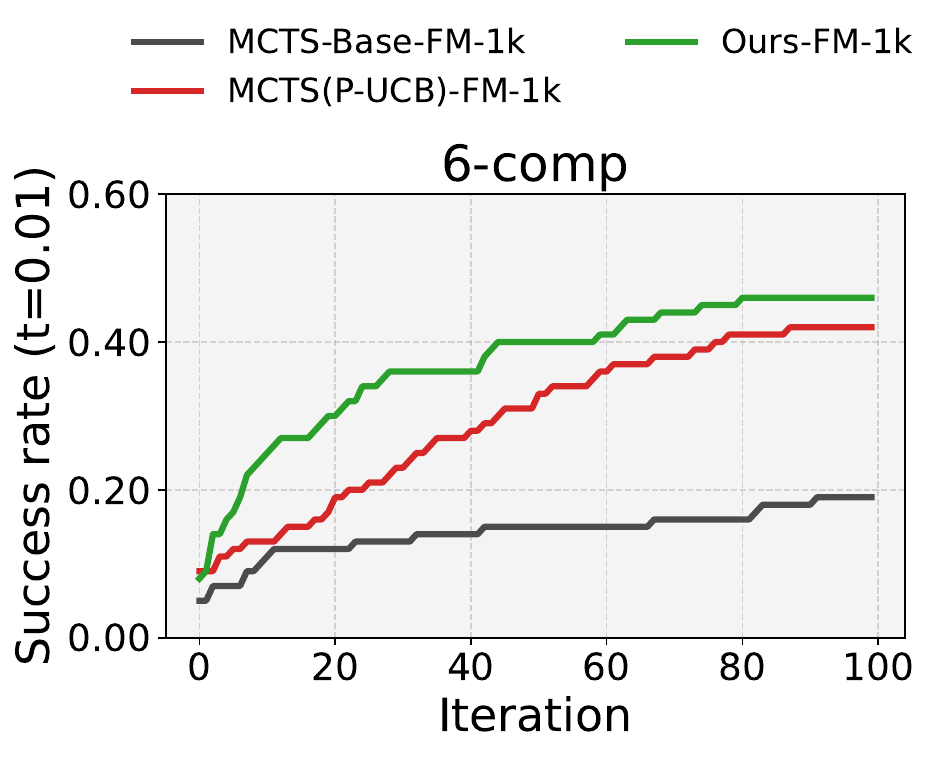} 
\caption{Success rate of our method with MCTS (Baseline) and its MCTS(P-UCB) variants.}
\label{fig:ablation}
\end{figure}

Within our tree structure framework, we incorporate probability guidance during node selection to leverage the LLM's capabilities in directing tree-based sampling. To validate the effectiveness of this LLM-provided probability guidance, we conducted a comparative analysis by modifying the MCTS baseline from standard UCB node selection to our version of P-UCB node selection. Figure~\ref{fig:ablation} confirms that the probability guidance from P-UCB is useful. However, our method still outperforms MCTS(P-UCB) through the implementation of node shrinking, which effectively addresses the challenges posed by structural tokens in circuit formulation. This advantage is particularly pronounced during the early exploration iterations.

\end{document}